\documentclass[lettersize,journal]{IEEEtran}
\usepackage{amsmath,amsfonts}
\usepackage{algorithmic}
\usepackage{algorithm}
\usepackage{array}
\usepackage[caption=false,font=normalsize,labelfont=sf,textfont=sf]{subfig}
\usepackage{textcomp}
\usepackage{stfloats}
\usepackage{url}
\usepackage{verbatim}
\usepackage{graphicx}
\usepackage{cite}
\usepackage{booktabs}
\usepackage{color}
\usepackage{multirow}
\usepackage{arydshln}
\begin{document}

\title{NeSLAM: Neural Implicit Mapping and Self-Supervised Feature Tracking With Depth Completion and Denoising }

\author{Tianchen~Deng,
        Yanbo~Wang,
        Hongle~Xie,
        Hesheng Wang,~\IEEEmembership{Senior Member,~IEEE},
        Jingchuan~Wang, 
        \\
        Danwei Wang,~\IEEEmembership{ Fellow,~IEEE},
        Weidong~Chen,~\IEEEmembership{Member,~IEEE}
        
\thanks{Tianchen Deng, Yanbo Wang, Hongle Xie, Jingchuan Wang, Hesheng Wang, Weidong Chen are with Institute of Medical Robotics and Department of Automation, Shanghai Jiao Tong University, and Key Laboratory of System Control and Information Processing, Ministry of Education, Shang hai 200240, China. Danwei Wang is with School of Electrical and Electronic Engineering, Nanyang Technological University, Singapore.
This work is supported by the National Key R\&D Program of China (Grant 2020YFC2007500), the National Natural Science Foundation of China (Grant U1813206), and the Science and Technology Commission of Shanghai Municipality (Grant 20DZ2220400). (*corresponding author:~wdchen@sjtu.edu.cn.)}
}
%


\maketitle

\begin{abstract}
   In recent years, there have been significant advancements in 3D reconstruction and dense RGB-D SLAM systems. One notable development is the application of Neural Radiance Fields (NeRF) in these systems, which utilizes implicit neural representation to encode 3D scenes. This extension of NeRF to SLAM has shown promising results. However, the depth images obtained from consumer-grade RGB-D sensors are often sparse and noisy, which poses significant challenges for 3D reconstruction and affects the accuracy of the representation of the scene geometry. Moreover, the original hierarchical feature grid with occupancy value is inaccurate for scene geometry representation. Furthermore, the existing methods select random pixels for camera tracking, which leads to inaccurate localization and is not robust in real-world indoor environments. To this end, we present NeSLAM, an advanced framework that achieves accurate and dense depth estimation, robust camera tracking, and realistic synthesis of novel views. First, a depth completion and denoising network is designed to provide dense geometry prior and guide the neural implicit representation optimization. Second, the occupancy scene representation is replaced with Signed Distance Field (SDF) hierarchical scene representation for high-quality reconstruction and view synthesis. Furthermore, we also propose a NeRF-based self-supervised feature tracking algorithm for robust real-time tracking. Experiments on various indoor datasets demonstrate the effectiveness and accuracy of the system in reconstruction, tracking quality, and novel view synthesis.
\end{abstract}


\section{Introduction}
\label{sec:intro}
Visual Simultaneous Localization and Mapping (SLAM) has made significant progress and has various applications in different fields, such as autonomous driving, indoor robotics, and virtual reality (VR). For real-world deployment, a system must possess several essential properties. Firstly, it should be capable of incrementally constructing an accurate geometric representation of the scene and estimating camera pose in real-time. Secondly, the system should demonstrate robustness in handling noisy and incomplete observations, while also being scalable to handle large-scale scenarios. Additionally, the ability to synthesize novel views can provide valuable benefits for applications such as virtual reality roaming.
\begin{figure}[t]
    \centering
    \includegraphics[width=\linewidth]{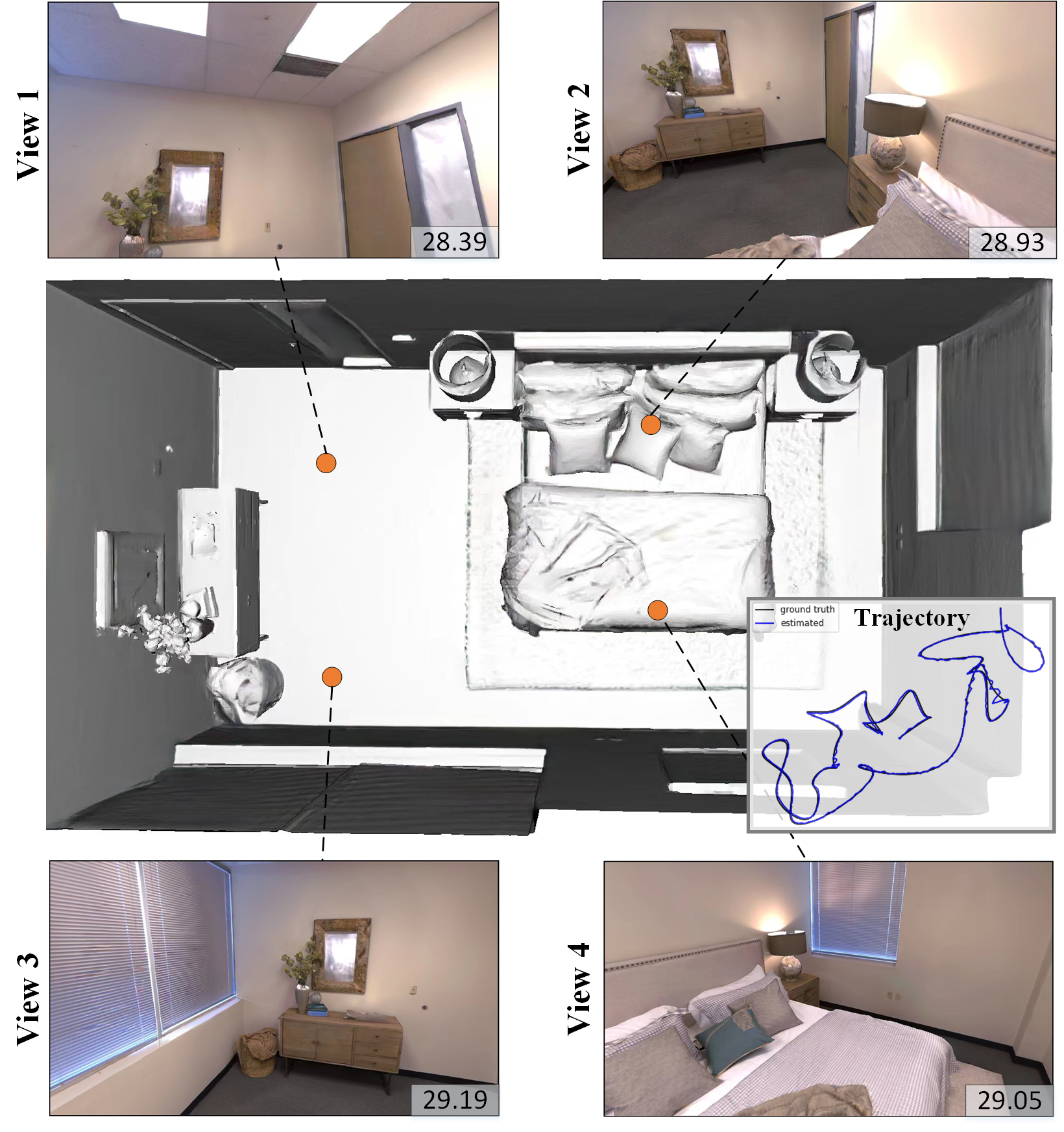}
    \vspace{-0.2cm}
    \caption{3D reconstruction and novel view synthesis results using NeSLAM. The final reconstruction mesh and images of novel view synthesis at different locations showcase the powerful scene reconstruction capability of our algorithm. We provide the PSNR value in the bottom right corner.}
    \label{fig:visualization}
    \vspace{-0.5cm}
\end{figure}

As for existing visual SLAM systems, there are several categories of them, such as sparse map points SLAM systems~\cite{orbslam2,deng,xie,xie2}, and dense SLAM systems~\cite{dtam,BADslam,codemapping}. Those systems are able to perform real-time pose estimation and can be employed in large-scale scenes with loop closing~\cite{loopclosing}. However, they fall short in terms of their scene representation capabilities. They tend to inadequately capture and incorporate essential information, resulting in incomplete and limited scene representations. Sparse map representation methods~\cite{liu1} are not suitable for subsequent tasks in robotics, such as navigation and obstacle avoidance~\cite{liu2,deng2}. With the rapid advances in deep learning, some learning-based SLAM systems are successively proposed to improve the ability of scene representation, such as Codeslam and Scenecode~\cite{codeslam,scenecode,droid}. 

 Compared with other representation methods, Neural radiance fields (NeRF) \cite{NeRF} is a promising recent advance technology with various application in robotics and autonomous driving~\cite{prosgnerf}. NeRF utilizes differrentiable rendering techniques and multi-layer perceptrons (MLP) to estimate the density and color of each point along a ray. The MLP has the ability to encode scene geometry in fine detail. Adopting these implicit representation methods in SLAM, there are several recently proposed systems such as iMAP~\cite{Sucar_2021_imap} and NICE-SLAM~\cite{Zhu_2022_niceslam}, and so on~\cite{plgslam,li1,li2,compact} . Both of them successfully combine NeRF with SLAM and achieve real-time pose estimation and dense mapping.

However, there are two key challenges for dense visual SLAM. The first challenge arises from the inherent limitations of consumer-grade RGB-D sensors, which result in sparse and noisy depth images. These characteristics pose a considerable obstacle to neural implicit mapping, as they heavily rely on accurate geometry information. The second challenge lies in the limitations of existing methods when it comes to tracking in real-world indoor scenes. These methods use random random pixel selection strategy, which often exhibit low tracking accuracy and are prone to failure. 

To this end, we propose NeSLAM, a dense RGB-D SLAM system that can represent the scene implicitly, camera tracking, and have the ability of novel view synthesis. For the first challenge, a depth completion and denoising network is proposed. This network aims to generate dense and precise depth images with depth uncertainty images. This geometry prior information plays a crucial role in guiding neural point sampling and optimizing the neural implicit representation. This network is used to improve the geometry representation capability and refine the performance of the entire system. 

For the second challenge, we propose a NeRF-based self-supervised feature tracking network specifically designed for accurate and real-time camera tracking in indoor scenes. This network leverages the strengths of NeRF with feature tracking to enable self-supervised optimization during the system operation, which can enhance the generalization capability. The keypoint network can better adapt to different complex scenes and make the system more stable, accurate, and scalable. We evaluate the effectiveness of the method on different indoor RGB-D datasets and do exhaustive evaluations and ablation experiments on these datasets. Our system demonstrates superior performance compared to recent and concurrent methods~\cite{Zhu_2022_niceslam,Sucar_2021_imap} that employ implicit mapping approaches.
\textbf{In summary, our contributions are shown as follows:}
\begin{itemize}
    \item A novel dense visual SLAM system is proposed with hierarchical implicit scene representation. This system is scalable, predictive, and robust to complex indoor scenes. It is an end-to-end, incrementally optimizable method for tracking and mapping. It offers the capability of generating photo-realistic novel views and producing accurate 3D meshes.
    \item A depth completion and denoising network is designed to provide dense and accurate depth images associated with depth uncertainty images. This geometry prior information is used to guide the point sampling process and improve geometric consistency. In addition, we replace the occupancy value with Signed Distance Field (SDF) value to better represent scene geometry.

    \item We propose a NeRF-based self-supervised feature tracking method for accurate and robust camera tracking in large and complex indoor environments, which is proven effectiveness and robust in our experiments.
\end{itemize}


\begin{figure*}[t]
  \centering
   \includegraphics[width=\linewidth]{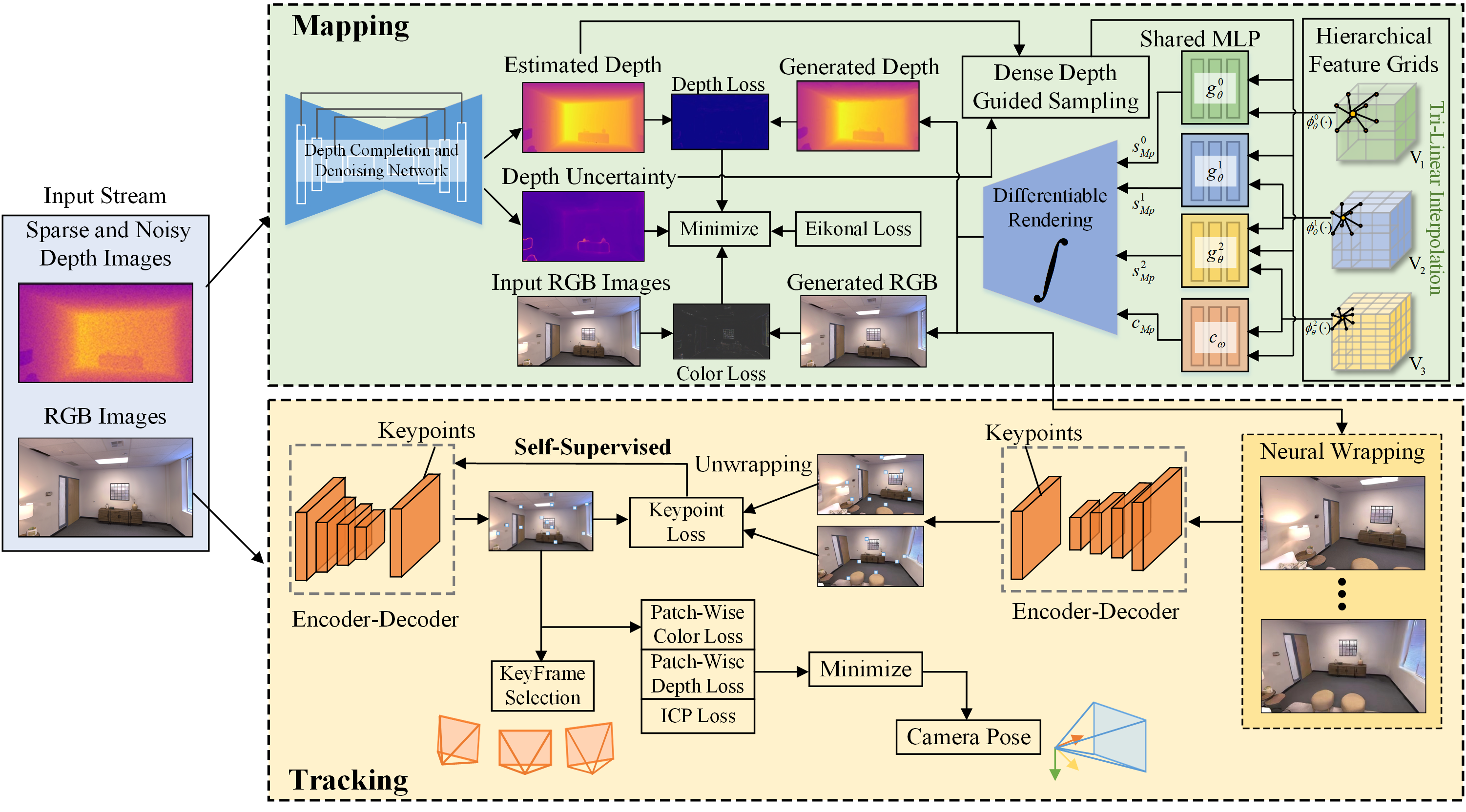}
    \vspace{-0.4cm}
\caption{The pipeline of our system. The input stream of our system is RGB and depth images, and the output is the implicit scene representation, generated RGB, depth images, depth uncertainty images, and the camera pose. Our system has two parallel threads: the mapping thread and the tracking thread. In the mapping thread, we estimate the dense and accurate depth image along with depth uncertainty. Then we use them to guide the neural point sampling and implicit representation optimization. The hierarchical feature grids are online updated by minimizing our carefully designed loss through differentiable rendering with the system operating. As for the tracking thread, we propose a NeRF-based self-supervised feature tracking network for accurate and robust pose estimation. This network is online self-supervised optimized via backpropagating keypoint loss. Those two threads are running with an alternating optimization.}
   \label{fig:pipeline}
   \vspace{-0.5cm}
\end{figure*}

 \vspace{-0.3cm}
\section{Related Work}
 \vspace{-0.1cm}
\label{sec:formatting}
\noindent \textbf{Visual SLAM System}\quad Traditional real-time visual SLAM systems depend on the constructed maps. PTAM~\cite{ptam}, a breaking SLAM work with parallel tracking and mapping, provides an effective method
for keyframe selection, feature matching, and camera localization for every frame. Some sparse mapping methods~\cite{orbslam2,vins,RAM}, which are then proposed that use manipulated keypoint for tracking, mapping, relocalization, and loop closing. 
These systems are robust to severe motion clutter and large indoor environments.

For learning-based SLAM systems, DTAM~\cite{dtam} is one of the pioneer works that use the dense map and view-centric scene representation. Some recent dense SLAM systems, such as~\cite{demon}, adopt the framework of DTAM to estimate pose and depth. Kinectfusion\cite{kinectfusion} explicitly represents the surface of the environments with a fixed resolution of volume, but it is costly in memory. Bundle-Fusion and Ba-net\cite{Bundle-fusion,ba-net} are dense SLAM systems that successfully use bundle adjustment for pose estimation. Other methods, such as CodeSLAM\cite{codeslam} propose a new compact but dense representation of scene geometry with a latent code. And \cite{dynamic} use the probabilistic field on the Lie group Sim(3) manifold for SLAM in a dynamic environment. In contrast to these methods, we use implicit scene mapping, which allows us to achieve more accurate geometry representation and novel view synthesis along the trajectory.

\noindent \textbf{Implicit Scene Representation} \quad Scene reconstruction has made significant progress recently\cite{reconstruct2,reconstruct3}. With the proposal of Neural radiance fields (NeRF)~\cite{NeRF}, many researchers explore to combine this implicit method into 3D reconstruction. NeRF is a ground-breaking method for novel view synthesis. It represents the scene with an MLP and renders images with the predicted volume densities along the rays. However, the representation of volume densities can not commit the geometric consistency, leading to poor surface prediction for reconstruction tasks. In order to deal with it, some methods are proposed that combine world-centric 3D geometry representation with neural radiance fields, such as UNISURF~\cite{UNISURF} and NeuS~\cite{NEUS}. UNISURF uses a unified way to formulate the implicit surface model with radiance fields. It enables more efficient points sampling and reconstructs accurate surfaces without input masks. NeuS~\cite{NEUS} replaces the volume density with Signed Distance Field (SDF) values. It proposes a new rendering formulation and incorporates additional depth measurements. Other methods~\cite{neural_rgbd,transformerfusion,VolumeFusion,neuralrecon, hierarchical} use various scene geometry representation methods, such as truncated signed distance function, voxel grid, or occupancy grid with latent codes. However, they all need ground-truth camera poses.

\noindent \textbf{NeRF with SLAM} \quad
Some works focus on pose estimation of NeRF, iNeRF~\cite{inerf}, NeRF--~\cite{nerf--} are concurrent work to estimate the camera pose with inverse NeRF optimization when the neural implicit network is fully trained. Without the pre-trained neural implicit network, BARF~\cite{barf} is proposed to train a neural network with inaccurate poses images or unknown poses images through bundle adjustment. However, their methods can not optimize poses and neural implicit network simultaneously. Pushing this to the limits, iMAP~\cite{Sucar_2021_imap}, and NICE-SLAM~\cite{Zhu_2022_niceslam} are successively proposed to combine neural implicit mapping with SLAM. iMAP uses a single multi-layer perceptron (MLP) to represent the scene, and NICE-SLAM uses a learnable hierarchical feature grid.
These are the works most relevant to our approach, but our method differs from them in the following ways. With the designed depth completion and denoising network, we can get more accurate reconstruction and novel view synthesis. We also propose a self-supervised feature tracking method for robust pose estimation in complex environments.

\vspace{-0.1cm}
\section{Method}
\subsection{System Overview}
The pipeline of our system is shown in Fig.~\ref{fig:pipeline}. Following the prior works~\cite{Sucar_2021_imap,Zhu_2022_niceslam}, we use three-level hierarchical feature grids and their corresponding decoders to represent the scene geometry. We also use another feature grid and corresponding decoder for color representation. For the implicit mapping thread, a depth completion and denoising network (Sec.~\ref{sec:depth_completion}) is designed to estimate dense depth images along with depth uncertainty images to strengthen geometry representation ability and improve sampling efficiency. Then the dense depth images and depth uncertainty are used to guide the neural point sampling and NeRF optimization. We also incorporate hierarchical neural scene representation with SDF into our system (Sec.~\ref{sec:representation}). For the camera tracking thread, a self-supervised feature tracking method (Sec.~\ref{sec:superpoint}) is designed for robust and accurate pose estimation. Several carefully designed loss functions are proposed to jointly optimize the scene implicit representation and camera pose estimation (Sec.~\ref{sec:optimization}). The network is incrementally online and updated with the system operation. 
\vspace{-0.4cm}
\subsection{Depth Completion and Denoising Network}
\vspace{-0.1cm}
\label{sec:depth_completion}
With the limitations of consumer-grade RGB-D cameras, the input depth images have two downsides. Firstly, the input depth images are relatively sparse because depth cameras often fail to sense depth for shiny, bright, transparent, and distant surfaces. Secondly, the input depth images are often noisy and have outliers, which is harmful for implicit geometry representation. In order to address those two downsides, we propose our depth completion and denoising network $D_{\theta}$ inspired by \cite{dense_depth,cspn}. The architecture of our depth network is shown in Fig. 3.
The input of our network is RGB images $I_i$, and sparse depth images ${D_i}$. The output of our network is initial depth prediction $D_{ini}$, non-local neighbor affinities $\omega_i$,  confidence map $\gamma_i$, and standard deviations $S_i$
\begin{equation}
\vspace{-0.1cm}
    D_{\theta}(I_i, D_i) = (D_{ini}, \omega_i,\gamma_i, S_i)
    \vspace{-0.1cm}
\end{equation}

\begin{figure*}[t]
  \centering
   \includegraphics[width=0.9\linewidth]{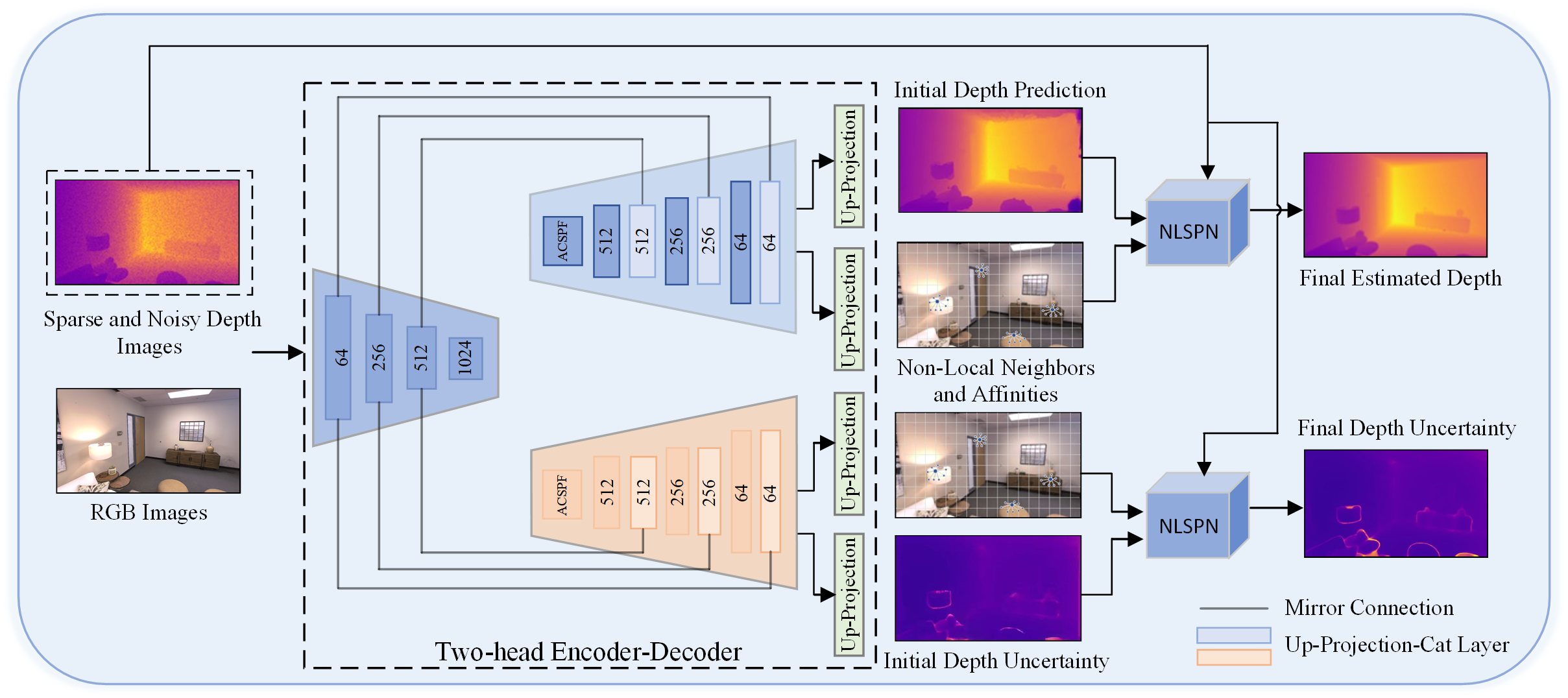}
   \label{fig:depth_net}
\caption{The architecture of our depth completion and denoising network. We use sparse and noisy depth images and corresponding RGB images as our input. We design a two-head encoder-decoder architecture to estimate dense depth along with depth uncertainty. We use mirror connections to add feature information from the encoder to the Up-Projection-Cat layer. The sparse depth map is embedded into the NLSPN module to guide the depth refinement.  }
\vspace{-0.3cm}
\end{figure*}
We use the residual network~\cite{resnet} as the backbone of our two-head encoder-decoder architecture, which is a UNet-like~\cite{Unet} neural network. We design a two branches architecture network with mirror connections to predict dense depth $D_i$ jointly with standard deviation $S_i$. The detailed parameters of the network are annotated in Fig. 3. In order to avoid the spatial information weaken with the down-sampling operation of the network, we add mirror connections by directly concatenating the feature from the encoder to decoder layers, which is the "Up-Projection-Cat" layer in Fig. 3. The feature dimensions of each layer of our encoder are 64, 256, 512, 1024. And the output feature dimensions of each
layer of the decoder are 512, 512, 256, 256, 64, and 64, respectively. We utilize an ACSPF module~\cite{cspn}, which combines Convolutional Spatial Pyramid Pooling (CSPP), Atrous Spatial Pyramid Pooling (ASPP), and Convolutional Feature Fusion (CFF) modules. The Up-Projection layer in Figure 3 is composed of the convolutional (conv) layer, batch normalization (bn) layer, and upsampling layer (bilinear interpolation). For better spatial information propagation, we employ Non-Local Spatial Propagation Network (NLSPN)~\cite{nlspn} to refine depth and depth uncertainty. This network uses non-local spatial propagation to estimate missing values and refine less confident values by propagating neighbor values with corresponding affinities. This refinement procedure makes the blurry depth images become more detailed. We also incorporate the confidence map ${\gamma_i}$ of the depth prediction to avoid negative influence from unreliable depth values during non-local propagation. This helps us get better results in depth completion and denoising. 

For network training, we train our model on Replica \cite{straub2019replica} dataset , Scannet \cite{scannet} dataset, and TUM RGB-D\cite{tum} dataset. Under the assumption that the depth and standard deviation are normally distributed, we use the negative log-likelihood of a Gaussian loss:
\begin{equation}
    L_{\theta} = \frac{1}{n}\sum_{i=1}^n(log(S_i^2)+\frac{(\hat{D}_i-D_{gt})^2}{S_i^2})
\end{equation}
where $D_i$, $S_i$ are the estimated depth and uncertainty of pixel i. n is the number of valid pixels in depth images.
\vspace{-0.3cm}
\subsection{Neural Scene Representation}
\vspace{-0.1cm}
\label{sec:representation}
\noindent \textbf{Scene Representation} \quad Following NeRF~\cite{NeRF} and NICE-SLAM~\cite{Zhu_2022_niceslam}, we incorporate hierarchical scene representation into our system. We use multi-level grid features with corresponding pre-trained MLPs for scene geometry representation. Inspired by VolSDF~\cite{VolSDF}, we change the occupancy with the Signed Distance Field (SDF) value which greatly improves the ability of geometry representation. For geometry representation, the feature grid is encoded into three levels: coarse $g_{\theta}^0(\cdot)$, middle $g_{\theta}^1(\cdot)$, fine $g_{\theta}^2(\cdot)$. With the corresponding geometry MLP decoder $g_{\theta}(\cdot)$, we can get the SDF value $s_{Mp}$ and geometry feature $z_{Mp}$ by querying the decoder.  For any map point $Mp \in \mathbb{R}^3$:  
\begin{gather}
\vspace{-0.1cm}
   coarse:s_{Mp}^0,z_{Mp}^0=g^0_{\theta}(Mp,\phi_{\theta}^0(Mp)) \notag \\ middle:s_{Mp}^1,z_{Mp}^1=g^1_{\theta}(Mp,\phi_{\theta}^1(Mp)) \notag\\
   fine:s_{Mp}^2,z_{Mp}^2=g^2_{\theta}(Mp,\phi_{\theta}^1(Mp),\phi_{\theta}^2(Mp))
   \vspace{-0.1cm}
\end{gather}
where $\theta$ is an optimizable parameter for feature grids. The optimization for geometry is a coarse-to-fine process. We first use the mid-level grid to represent the coarse-level scene geometry and use the fine-level grid for refinement. For the coarse and mid-level grid, the features are directly decoded into SDF values and features with corresponding MLPs. For the fine-level grid, it is a residual value of the mid-level grid. We concatenate the mid-level feature $\phi_{\theta}^1(Mp)$ with the fine-level feature $\phi_{\theta}^2(Mp)$ as the input of the fine-level decoder. The output of the fine-level decoder
is an offset from mid-level SDF value. The final SDF value $\hat{s}$ is defined as:
\begin{equation}
    \hat{s}=s_{Mp}^1+s_{Mp}^2
\end{equation}

In our framework, these three pre-trained decoders are fixed for optimization stabilization and geometric consistency. We only optimize the feature grids $g_{\theta}(\cdot)$ during the optimization process. The coarse-level feature grid is primarily used to extract low-frequency information (such as contours), while the fine-level feature grid is used to extract high-frequency information (such as detailed texture features) from the environment.

 For color representation, we use another feature grid $\varphi_{\omega}$ and decoder $c_{\omega}$: 
 \begin{equation}
     color: c_{Mp}=c_{\omega}(Mp,z_{Mp}^0,z_{Mp}^1,z_{Mp}^2, \varphi_{\omega}(Mp))
 \end{equation}
where $\omega$ is the learnable parameter of the color feature grid. During the optimization process, we jointly optimize the feature grids $\varphi_{\omega}(Mp)$ and decoder $c_{\omega}$ for global color consistency and incrementally learning. With the prior of \cite{Zhu_2022_niceslam}, the feature dimension is 64 and 5 layers for the geometry and two layers for color decoders. We also incorporate the Gaussian positional encoding \cite{fourier,Zhu_2022_niceslam} to $Mp$ for better learning of high-frequency details of both color and geometry. 

\noindent \textbf{Differentiable Rendering} \quad Following NeRF \cite{NeRF}, we use the predicted SDF value and colors from decoders and integrate them for scene representation. We can determine a ray $r(t)=o+td$ whose origin is at the camera center of projection $o$. We sample points along this ray. The sample bound is within the near and far planes $t_k\in [t_n,t_f]$, $k \in {1,\dots,K}$. For every sample point $Mp_k$, we can get three level SDF values and color of them. We follow VolSDF~\cite{VolSDF} to transform the SDF value into density value:
\begin{equation}
    \sigma_\beta(s)= \begin{cases}\frac{1}{2 \beta} \exp (\frac{s}{\beta}) & \text { if } s \leq 0 \\ \frac{1}{\beta}\left(1-\frac{1}{2} \exp (-\frac{s}{\beta})\right) & \text { if } s>0\end{cases}
\end{equation}
where $\beta \in \mathbb{R} $ is a learnable parameter that controls the sharpness of the surface boundary.
Then we define the termination probability as:
\begin{gather}
    coarse:\omega_k^c=\prod_{j=1}^{k-1}\exp \left(- \sigma_j^c \delta_j\right)\left(1-\exp \left(-\sigma_k^c\right)\right) \notag \\ 
    fine:\omega_k^f=\prod_{j=1}^{k-1}\exp (- \sigma_j^f \delta_j)(1-\exp (-\sigma_k^f)) 
\end{gather}
\begin{figure*}[t]
    \centering
    \includegraphics[width=\linewidth]{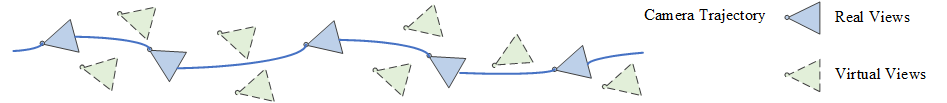}
    \caption{The generation of virtual images.  We synthesize novel view images with different poses $\{R,T\}$ for every coming keyframe. we generate its
RGB and depth images and filter out inaccurate depths through a geometric consistency check. }
    \label{fig:virtual}
\end{figure*}

where $\delta_j$ represents the distance between neighboring sample points.
Then the color, depth, and standard deviation $D_s$ of the ray are computed from the rendering weights $\omega_k$:
\begin{gather}
    \hat{C} = \sum_{k=1}^K\omega_k^fc_k \quad
    \hat{D}^f = \sum_{k=1}^K\omega_k^ft_k\quad
    \hat{D}^c = \sum_{k=1}^k\omega_k^ct_k \notag \\ 
    \hat{S^f}^2 = \sum_{k=1}^K\omega_k^f(D_f-t_k)^2 \quad 
    \hat{S^c}^2 = \sum_{k=1}^K\omega_k^c(D_c-t_k)^2
\end{gather}

\noindent \textbf{Depth Guided Sampling} \quad
The estimated depth images and depth uncertainty provide valuable geometry information which can guide neural point sampling along a ray within the bounds of depth uncertainty. For a room-scale scene in the Replica dataset, we get $N_{strat}$ points for stratified sampling between the near and far planes. Then, $N_{surface}$ points are drawn from the Gaussian distribution determined by the depth prior $\mathcal{N} (D,S^2)$. When the depth is not known or invalid, we use the estimated depth prior from differentiable rendering and sample points according to $\mathcal{N} (\hat{D^f},\hat{S^f}^2)$. Compared to the original methods, such as NeRF or NICE-SLAM, our approach allows for more efficient point sampling and enhances the scene representation capability of the network.  
\vspace{-0.2cm}

\subsection{NeRF-Based Self-Supervised Feature Tracking}
\label{sec:superpoint}
We parallel run this thread for pose estimation in real-time e.g.the rotation and translation $\{R,T\}$. In prior work \cite{Zhu_2022_niceslam}, they random sample $P_t$ pixels in the current frame to optimize the camera pose. However, random sampling is not fit for large scenes and noising observations, which are really common in real-world environments. The accuracy of their method is low, the robustness is poor, and the efficiency is also low. They fail in many situations, such as quick camera movement and large scenes. We consider that the keypoint is more suitable due to its inherent properties of rotation and translation invariance. To this end, we propose a Nerf-based self-supervised feature tracking network and incorporate it into our camera tracking thread. It can self-supervised optimize during the system operation compared with a superpoint network and achieve high localization accuracy in different scenarios.

\noindent \textbf{Network Architecture} \quad With the prior work~\cite{superpoint}, we use a fully-convolutional neural network architecture. The input is full-sized images, and the output is keypoints detections. We use a VGG-style encoder to reduce the dimensionality of the image. The encoder maps the input image $I \in \mathbb{R}^{H\times W}$ to a feature map $\mathcal{F} \in \mathbb{R}^{H_f\times W_f\times \mathcal{B}}$, where $H_f=H/8$ and $W_f=H/8$. 

For the keypoint decoder, it uses $\mathcal{X}\in \mathbb{R}^{H_f \times H_f \times 65}$ tensor as input. We use 65 channels which contain $8\times8$ grid regions of pixels and an extra dustbin for no interest point area. Then we use a channel-wise softmax and remove the dustbin channel after that. The output is reshaped as $\mathbb{R}^{H\times W}$. 

\noindent \textbf{NeRF-Based Self-Supervised Refinement} \quad
The original superpoint model~\cite{superpoint} is trained on the MS-COCO image dataset\cite{coco} with homographic adaptation for domain adaptation. However, the pre-trained superpoint model is unsuitable for different real-world datasets. We want to incrementally optimize the superpoint model with the operating of our system. So we propose a NeRF-based self-supervised refinement method to achieve this.  
\begin{figure*}[t]
    \centering
    \includegraphics[width=\linewidth]{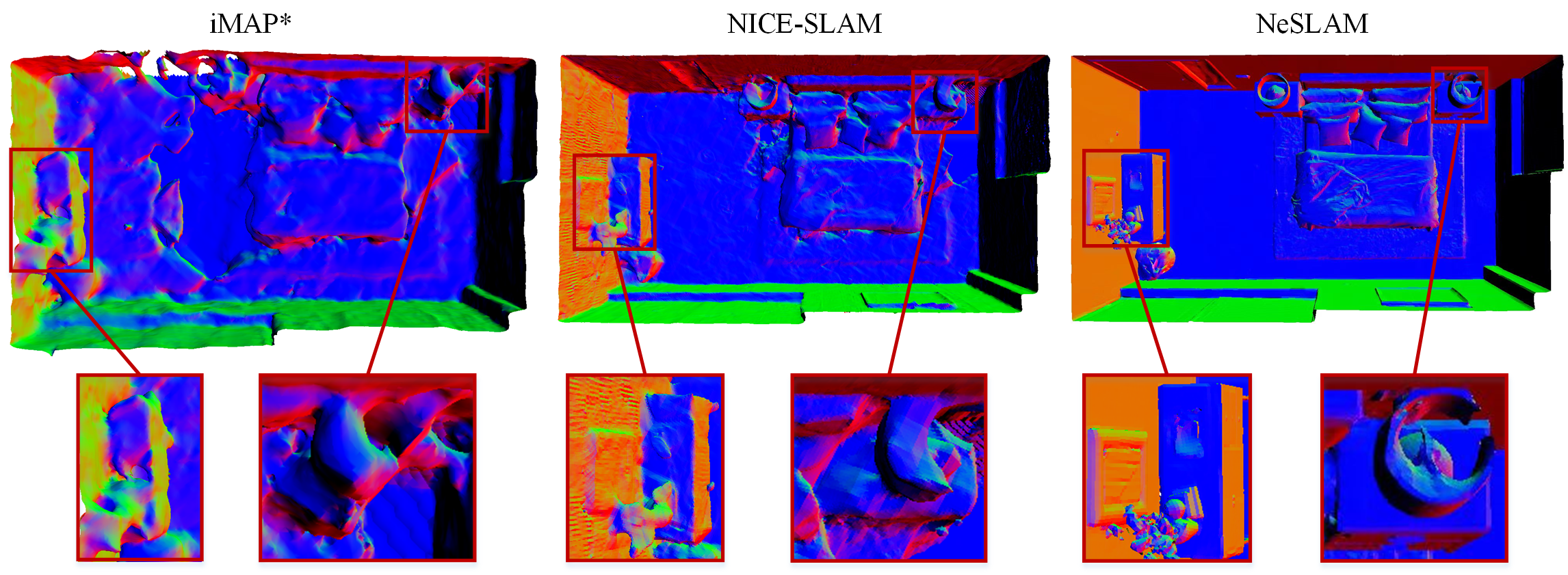}
    \vspace{-0.1cm}
    \caption{Qualitative reconstruction results on the Replica dataset\cite{straub2019replica} of room 1. From left to right, we show the construction meshes of iMAP*\cite{Sucar_2021_imap}, NICE-SLAM\cite{Zhu_2022_niceslam}, and our method. The red box highlights the improvements of our algorithm compared to other algorithms.}
    \label{fig:mesh_vis}
\vspace{-0.4cm}
\end{figure*}
\renewcommand\arraystretch{1.3}
\begin{table*}[]
\centering
\caption{Reconstruction Results on the Replica Dataset~\cite{straub2019replica}. We use three different metrics Acc.$\downarrow $, Comp.$\downarrow $, Comp. Ratio$\uparrow$. iMAP* refer 
to the re-implementation of iMAP provided in \cite{Zhu_2022_niceslam}.}
\begin{tabular}{@{}ccccccccccc@{}}
\toprule
Methods                                       & Metrics                           & room-0         & room-1         & room-2         & office-0       & office-1       & office-2       & office-3       & office-4       & Avg.           \\ \midrule
\multirow{3}{*}{iMAP*\cite{Sucar_2021_imap}}       & Acc.{[}cm{]} $\downarrow $                      & 3.28           & 3.49           & 4.48           & 5.57           & 3.41           & 4.72           & 4.09           & 4.61           & 4.21           \\
                                              & Comp.{[}cm{]} $\downarrow $                     & 4.96           & 4.74           & 5.31           & 6.01           & 5.13           & 5.51           & 5.29           & 6.47           & 5.43           \\
                                              & Comp. Ratio{[}\textless{}5cm \%{]} $\uparrow$ & 82.73          & 82.16          & 74.43          & 76.53          & 78.84          & 75.03          & 76.14          & 75.83          & 77.72          \\ \midrule
\multirow{3}{*}{NICE-SLAM\cite{Zhu_2022_niceslam}} & Acc.{[}cm{]} $\downarrow $                      & 2.93           & 2.97           & 3.03           & 4.86           & 2.95           & 3.71           & 3.04           & 2.65           & 3.27           \\
                                              & Comp.{[}cm{]} $\downarrow $                     & 2.95           & 2.92           & 2.87           & 3.95           & 3.63           & 3.24           & 3.51           & 3.65           & 3.34           \\
                                              & Comp. Ratio{[}\textless{}5cm \%{]} $\uparrow$ & 91.55          & 87.25          & 94.03          & 86.04          & 87.83          & 87.35          & 87.05          & 89.58          & 88.83          \\ \midrule
\multirow{3}{*}{NeSLAM}                        & Acc.{[}cm{]} $\downarrow $                      & \textbf{2.55}  & \textbf{2.11}  & \textbf{2.14}  & \textbf{2.13}  & \textbf{3.02}  & \textbf{3.23}  & \textbf{2.91}  & \textbf{2.45}  & \textbf{2.57}  \\
                                              & Comp.{[}cm{]} $\downarrow $                     & \textbf{2.32}  & \textbf{2.31}  & \textbf{2.27}  & \textbf{1.64}  & \textbf{1.67}  & \textbf{2.93}  & \textbf{3.03}  & \textbf{3.55}  & \textbf{2.46}  \\
                                              & Comp. Ratio{[}\textless{}5cm \%{]} $\uparrow$ & \textbf{91.78} & \textbf{94.67} & \textbf{91.97} & \textbf{95.55} & \textbf{94.56} & \textbf{90.91} & \textbf{90.49} & \textbf{91.32} & \textbf{92.66} \\ \bottomrule
\end{tabular}
\vspace{-0.4cm}
\label{tab:reconstruct}
\end{table*}
In our self-supervised approach, we use the pre-trained model for the base interest point detector. Then, we propose a novel neural wrapping procedure to get some different views of images for data augmentation. The generation of virtual images is shown in Fig. \ref{fig:virtual}. In this procedure, we synthesize novel view images with different poses $\{R,T\}$ for every coming keyframe. We input these images into the network to get the interest points detections. Then we unwrap these images into the initial pose and calculate the keypoint $L_p$ loss. The unwrapping procedure can be formulated as follows:
\begin{equation}
    \mathcal{X'}=\mathcal{X}_{\{R,t\}} \left\langle \Pi \left(D, \{R,T\}, K\right)\right\rangle
\end{equation}
where $\mathcal{X},\mathcal{X'}$ are the current keyframe and the corresponding unwrapped synthesis images. $D$ is the depth image and K is the camera intrinsics. Operator $\Pi()$ is the resulting 2D coordinates of projection. $\left\langle \right\rangle$ is the sampling operator. Then we can calculate keypoint loss:
\begin{align}
    L_p = \frac{1}{H_c W_c}\sum_{h=1,w=1}^{H_c,W_c}l_p(x_{hw};x'_{hw})
\end{align}
where $l_p$:
\begin{equation}
    l_p\left(\mathbf{x}_{h w} ; \mathbf{x'}_{h w}\right)=-\log \left(\frac{\exp \left(\mathbf{x}_{h w x'}\right)}{\sum_{k=1}^{65} \exp \left(\mathbf{x}_{h w k}\right)}\right)
\end{equation}
$l_p$ is cross-entropy loss over the cells $x_{hw}\in \mathcal{X}$, $x'_{hw} \in \mathcal{X'}$ from the current keyframe and the corresponding unwrapped synthesis images.

\subsection{Optimization in Mapping and Tracking}
\label{sec:optimization}
\vspace{-0.1cm}
In this section, we provide more details of the optimization of scene geometry $\theta$, color $\omega$, and camera poses $\{R,T\}$.

To optimize the scene feature grid in Section \ref{sec:representation}, we uniformly sample $P_t$ pixels from the current frame and the selected keyframes. Then, we iteratively optimize the feature grid to minimize the depth and color loss. The depth loss is defined as:
\begin{equation}
\vspace{-0.1cm}
    L_D^{l} = \frac{1}{P_t} \sum_{i=1}^{P_t} (log(\hat{S_i^l}^2)+\frac{\hat{D_i^l}-D_i^l}{\hat{S_i^l}^2})
\end{equation}
Here $D_i^l$ and $S_i^l$ are the target depth and standard deviation, $l\in {c,f}$. $\hat{D_i^l}$ and $\hat{S_i^l}$ are the estimated depth and standard deviation. We apply this loss to the pixel where one of the following conditions is true: 1) $|\hat{D_i^l}-D_i^l|>S_i^l$, the distance between generated depth and input depth is greater than the standard deviation value. 2) $\hat{S_i^l}>S_i^l$, the generated depth standard deviation. We also use this loss to optimize the pre-trained depth denoising network incrementally.
The color loss is defined as:
\begin{equation}
\vspace{-0.1cm}
    L_c = \frac{1}{P}\sum_{i=1}^P\left \| \hat{C_i}-C_i \right \|_1
    \label{equ:photometric}
    \vspace{-0.1cm}
\end{equation}

{
\setlength{\parindent}{0cm} where $\hat{C_i}$ and $C_i$ are the estimated color and target color. Inspired by NICE-SLAM, we use geometry loss to optimize mid-level feature at the first stage. Then we also use $L_D^l$ to jointly optimize mid and fine level feature. In addition, we add the Eikonal loss~\cite{Eikonal} to regularize the output SDF values:}
\begin{equation}
    \mathcal{L}_{\text {eikonal }}=\sum_{\mathbf{x} \in \mathcal{X}}\left(\|\nabla \hat{s}(\mathbf{x})\|_2-1\right)^2
\end{equation}

\begin{figure*}[t]
    \centering
    \includegraphics[width=0.95\linewidth]{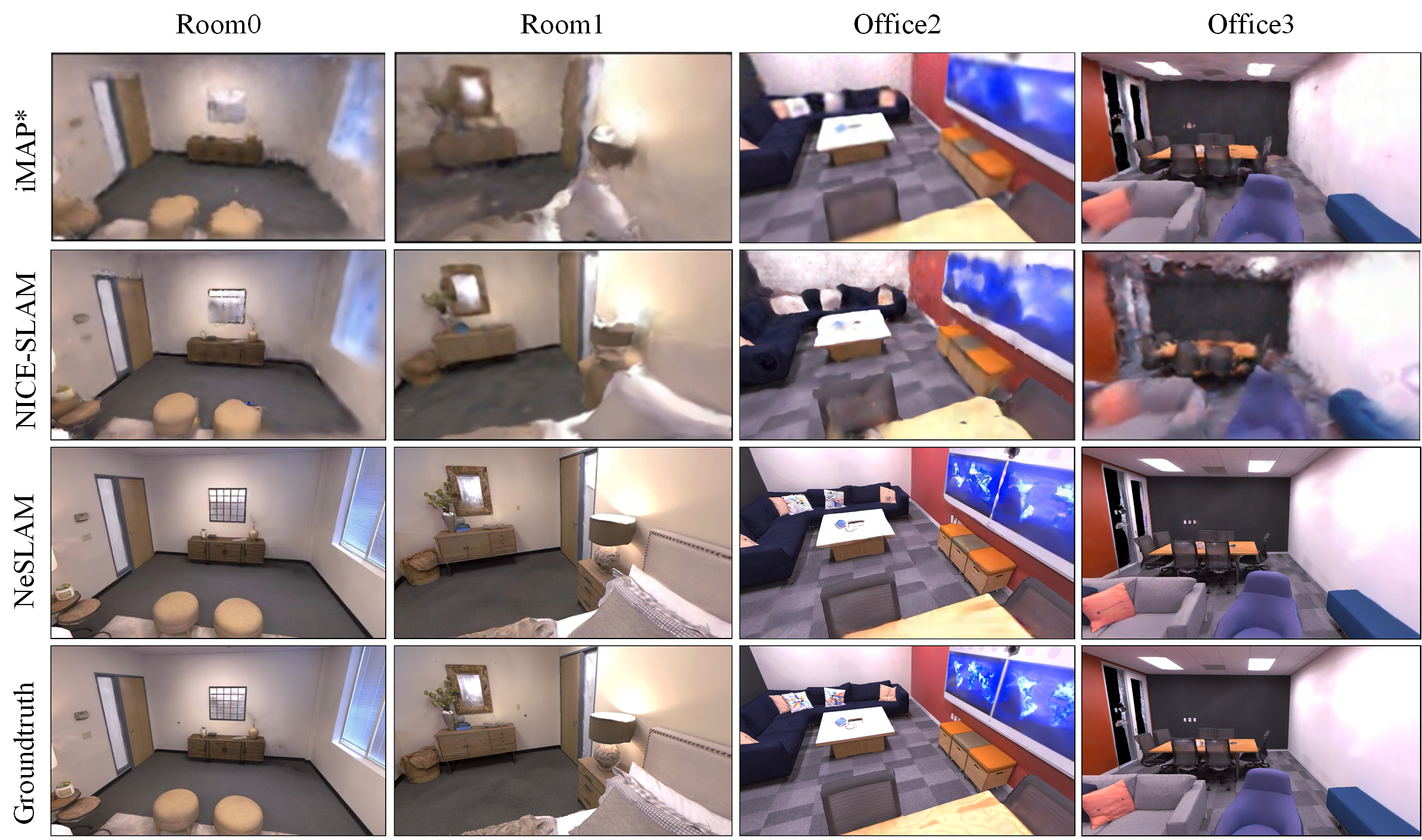}
    \caption{Qualitative results on the Replica dataset\cite{straub2019replica}. We show the view synthesis results of iMAP*\cite{Sucar_2021_imap}, NICE-SLAM\cite{Zhu_2022_niceslam}, and our method. Our method performs better than other methods with higher-quality view synthesis results.  }
    \label{fig:reconstruct}
    \vspace{-0.4cm}
\end{figure*}
\begin{figure}[t]
    \centering
    \includegraphics[width=0.99\linewidth]{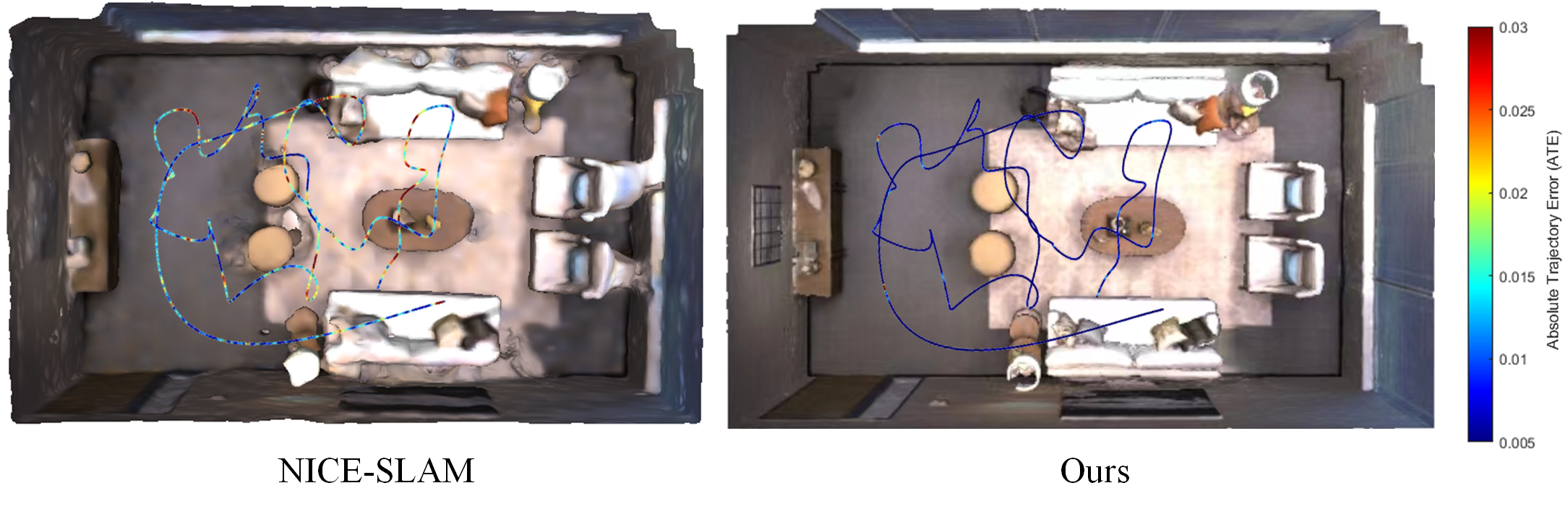}
    \caption{We depict the final mesh and camera tracking trajectory error (Absolute Trajectory Error) of different methods in replica dataset \cite{straub2019replica}.  The color bar on the right shows the ATE value.}
    \label{fig:comparison}
    \vspace{-0.5cm}
\end{figure}
\begin{table*}[]
\centering
\caption{Camera tracking Results on the Replica Dataset~\cite{straub2019replica} of traditional methods and learning-based methods.}
\vspace{-0.1cm}
\setlength{\tabcolsep}{2mm}{
\begin{tabular}{lcccccccccc}
\toprule
Methods                           & Metrics     & room-0          & room-1          & room-2          & office-0        & office-1        & office-2        & office-3        & office-4        & Avg.            \\ \midrule
\multirow{2}{*}{iMAP}             & RMSE{[}m{]} $\downarrow$ & 0.0553          & 0.0459          & 0.0239          & 0.0247          & 0.0177          & 0.0495          & 0.0697          & 0.0267          & 0.0391          \\
                                  & Mean{[}m{]} $\downarrow$ & 0.0345          & 0.0407          & 0.0206          & 0.0178          & 0.0165          & 0.0327          & 0.0591          & 0.0229          & 0.0306          \\ \midrule
\multirow{2}{*}{NICE-SLAM}        & RMSE{[}m{]} $\downarrow$ & 0.0225          & 0.0238          & 0.0199          & 0.0148          & 0.0128          & 0.0198          & 0.0223          & 0.0235          & 0.0199          \\
                                  & Mean{[}m{]} $\downarrow$ & 0.0191          & 0.0207          & 0.0156          & 0.0113          & 0.0107          & 0.0157          & 0.0185          & 0.0188          & 0.0163          \\ \midrule
\multirow{2}{*}{NeSLAM}           & RMSE{[}m{]} $\downarrow$ & 0.0060          & 0.0093          & \textbf{0.0052} & \textbf{0.0041} & 0.0043          & \textbf{0.0057} & 0.0096          & \textbf{0.0083} & 0.0066          \\
                                  & Mean{[}m{]} $\downarrow$ & 0.0053          & 0.0082          & \textbf{0.0045} & \textbf{0.0037} & 0.0038          & \textbf{0.0045} & 0.0076          & \textbf{0.0065} & 0.0056          \\ \cdashline{1-11}
\multirow{2}{*}{ORB-SLAM2(RGB)}   & RMSE{[}m{]} $\downarrow$ & 0.0050          & 0.0043          & 0.0225          & 0.0049          & 0.0048          & 0.1225          & 0.0077          & 0.1137          & 0.0356          \\
                                  & Mean{[}m{]} $\downarrow$ & 0.0044          & 0.0038          & 0.0199          & 0.0037          & 0.0041          & 0.1102          & 0.0065          & 0.0938          & 0.0308          \\ \midrule
\multirow{2}{*}{ORB-SLAM2(RGB-D)} & RMSE{[}m{]} $\downarrow$ & \textbf{0.0034} & \textbf{0.0027} & 0.0057          & 0.0048          & \textbf{0.0039} & 0.0058          & \textbf{0.0087} & 0.0098          & \textbf{0.0055} \\
                                  & Mean{[}m{]} $\downarrow$ & \textbf{0.0030} & \textbf{0.0021} & 0.0051          & 0.0039          & \textbf{0.0032} & 0.0048          & \textbf{0.0071} & 0.0085          & \textbf{0.0047} \\ \bottomrule
\end{tabular}}
\label{tab:tracking}
\vspace{-0.5cm}
\end{table*}

\begin{figure}[t]
    \centering
    \includegraphics[width=0.95\linewidth]{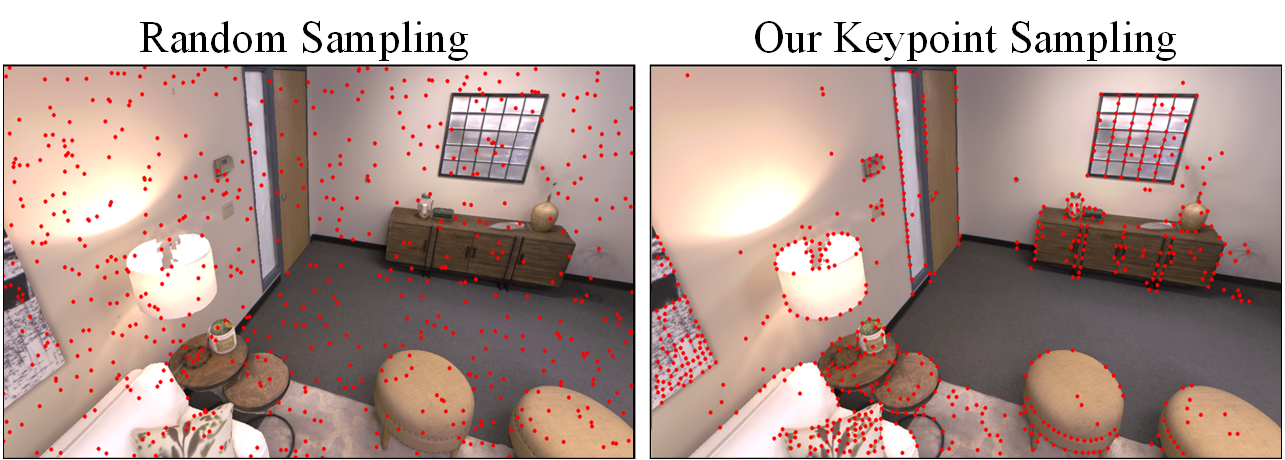}
    \caption{Qualitative results of our self-supervised feature tracking network on the Replica dataset. We show the sampling points of different methods.  }
    \label{fig:reconstruct6}
    \vspace{-0.5cm}
\end{figure}
where $\mathcal{X}$ are a set of uniformly sampled near-surface points.
Finally, we jointly optimize all level feature grids and the color decoder with the loss:
\begin{equation}
    \underset{\theta,\omega}{min}(L_D^f+L_D^c+\lambda_c L_c+\lambda_e L_{eikonal})
\end{equation}
This multi-stage optimization can lead to better geometry consistency and convergency. 

\noindent \textbf{Camera Tracking} \quad We use the extracted keypoints in the current frame to optimize the camera pose. We apply color loss in Eq.(\ref{equ:photometric}), and modified depth loss:
\begin{equation}
    L_{D\_v}=\frac{1}{P_t}\sum_{i=1}^{P_t}\frac{\left \| D_i-\hat{D}_i^c \right \|_1 }{\hat{S}^c}+\frac{\left \| D_i-\hat{D}_i^f \right \|_1 }{\hat{S}^f}
\end{equation}

This depth modified loss avoid less certain regions make influence on the reconstructed geometry. \\
\noindent \textbf{Patch-Wise Loss} \quad Furthermore, we replace original depth and color loss with patch-wise depth variance 
 loss $L_{p\_D\_v}$, patch-wise color loss $L_{p\_c}$, and patch-wise depth loss $L_{p\_D}$. We use $3 \times 3$ patch for every interest point to obtain better gradient descent and convergency. 
Finally, we incorporate ICP loss into our system to explicitly express the camera pose in loss function.
\begin{equation}
\vspace{-0.1cm}
    L_{ICP}=\sum_{i=1}^{P_t}\left \|\mathcal{C}(X_a^i-\Pi_{R,T}(X_b)) \right \| _1
\vspace{-0.1cm}
\end{equation}
where $X_a^i$ is ith keypoint of frame a, $X_b$ are the keypoints from frame b. Then, we project the keypoint from frame b into frame a and find the closest matching $\mathcal{C}$ of those keypoints. The ICP loss is defined as the pixel coordinate loss of the matching keypoints. 
The final tracking loss is defined as: 
\begin{equation}
    \underset{R,T}{min} (L_{p\_d\_v}+\lambda _{p\_D}L_{p\_D}+\lambda_{1}L_{p\_c}+\lambda_2 L_{ICP})
\end{equation}
we formulate this minimization problem to optimize camera poses.

\begin{figure}[t]
    \centering
    \includegraphics[width=0.95\linewidth]{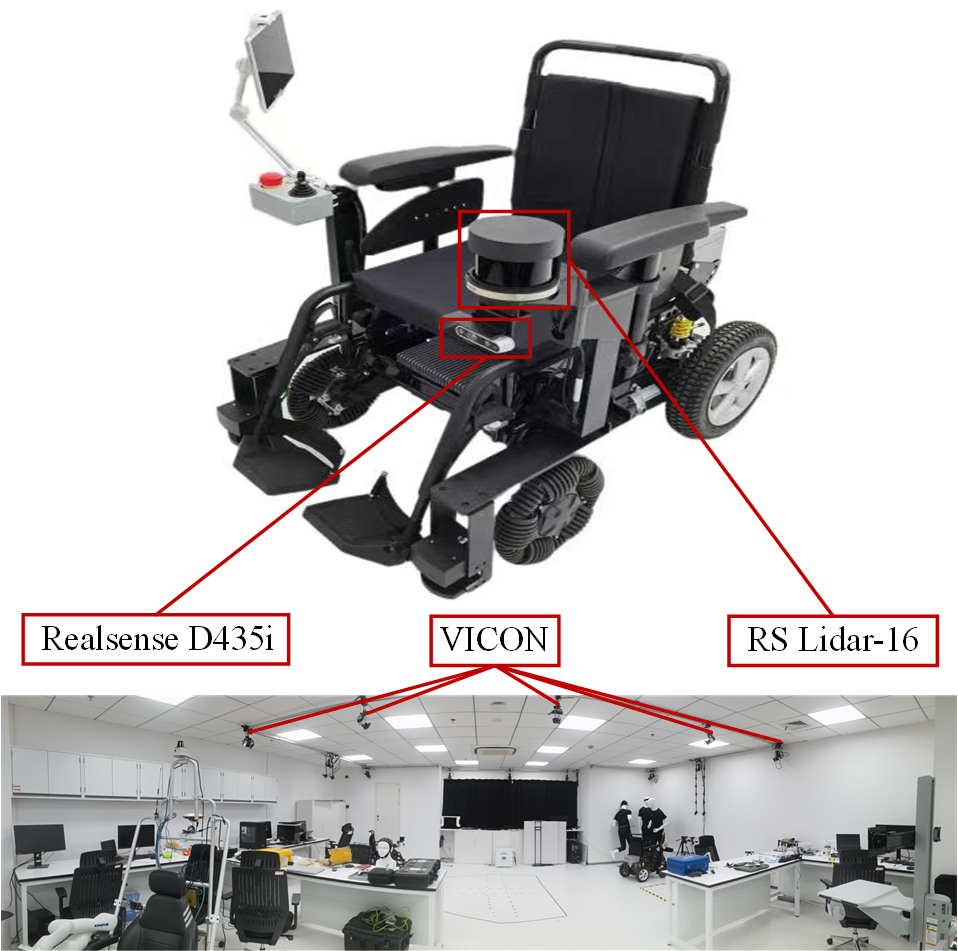}
    \caption{This is our wheelchair prototype, serving as our data collection platform. We also present a panoramic image of the indoor scenario. We use Realsense D435i to collect color and depth images and use VICON as groundtruth.}
    \label{fig:reconstruct2}
    \vspace{-0.4cm}
\end{figure}

\begin{table*}[!t]
\centering
\caption{Geometric (Depth L1) and Photometric (PSNR) results on the Replica\cite{straub2019replica} datasets. iMAP* refers to the re-implementation of iMAP provided in \cite{Zhu_2022_niceslam}.}
\vspace{-0.2cm}
\begin{tabular}{@{}ccccccccccc@{}}
\toprule
Methods                    & Metrics           & room-0         & room-1         & room-2         & office-0       & office-1       & office-2       & office-3       & office-4       & Avg.           \\ \midrule 
\multirow{2}{*}{iMAP*\cite{Sucar_2021_imap}}     & Depth L1 {[}cm{]} $\downarrow$ & 5.80           & 5.27           & 5.67           & 7.49           & 11.87          & 8.22           & 7.74           & 6.12           & 7.27           \\
                           & PSNR {[}db{]} $\uparrow$    & 20.17          & 20.37          & 19.98          & 24.37          & 23.01          & 18.07          & 24.03          & 21.55          & 21.44          \\ \midrule
\multirow{2}{*}{NICE-SLAM\cite{Zhu_2022_niceslam}} & Depth L1 {[}cm{]} $\downarrow$ & 1.81           & 1.44           & 2.04           & 1.39           & 1.76           & 8.33           & 4.99           & 2.01           & 2.97           \\
                           & PSNR {[}db{]} $\uparrow$     & 24.31          & 22.52          & 21.07          & 26.93          & 28.79          & 20.45          & 25.07          & 22.37          & 23.93          \\ \midrule
\multirow{2}{*}{NeSLAM}     & Depth L1 {[}cm{]} $\downarrow$ & \textbf{1.25}  & \textbf{2.01}  & \textbf{1.67}  & \textbf{1.02}  & \textbf{0.91}  & \textbf{4.02}  & \textbf{2.81}  & \textbf{1.53}  & \textbf{1.90}  \\
                           & PSNR {[}db{]} $\uparrow$     & \textbf{27.72} & \textbf{25.37} & \textbf{24.56} & \textbf{27.19} & \textbf{30.37} & \textbf{27.28} & \textbf{27.22} & \textbf{26.56} & \textbf{27.03} \\ \bottomrule
\end{tabular}
\label{tab:photometric}
\vspace{-0.4cm}
\end{table*}
\begin{figure}[!t]
    \centering
    \includegraphics[width=0.9\linewidth]{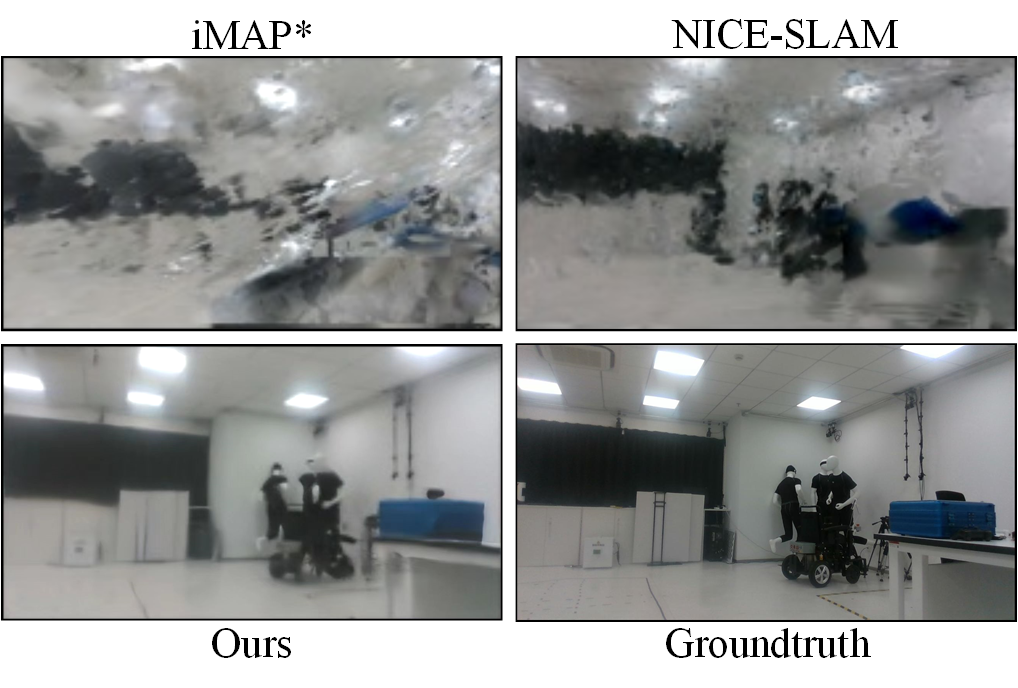}
    \vspace{-0.2cm}
    \caption{Qualitative results on our own real-world datasets. We show the view synthesis results of iMAP*\cite{Sucar_2021_imap}, NICE-SLAM\cite{Zhu_2022_niceslam}, and our method in scenes with fast camera movements and noisy input.  }
    \label{fig:reconstruct3}
    \vspace{-0.3cm}
\end{figure}
\begin{figure}[!t]
    \centering
    \includegraphics[width=0.9\linewidth]{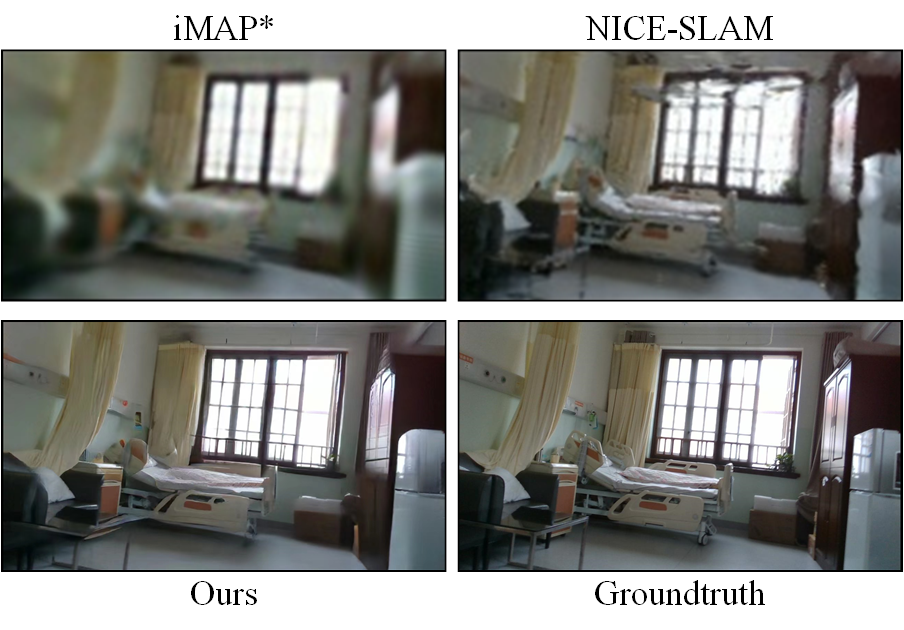}
    \vspace{-0.2cm}
    \caption{Qualitative results on our own real-world datasets (hospital). We show the view synthesis results of iMAP*\cite{Sucar_2021_imap}, NICE-SLAM\cite{Zhu_2022_niceslam}, and our method in real-world robot operational scenario.   }
    \label{fig:reconstruct4}
    \vspace{-0.6cm}
\end{figure}
\section{Experiments}
\subsection{Implementation Details}
We evaluate our method on various datasets and conduct a comprehensive ablation study to verify the effectiveness of our design. All training and evaluation experiments are conducted on a single NVIDIA RTX 3090 GPU. In all our experiments, we use $N_{strat}$ = 32 and $N_{surface}$ = 16 sampling points on a ray. The color loss weighting is $\lambda_c = 0.3$ and $\lambda_e = 0.1$ for mapping and $\lambda_1 = 0.15$ for tracking. The ICP loss weighting is $\lambda_2=0.2$ and patch-wise depth loss is $\lambda_{p\_D}=0.35$. For small-scale datasets, such as Replica, we select five active frames for mapping. For large-scale real-world datasets, we select ten active frames for mapping. we select Adam optimizer\cite{adam} ($\beta=(0.9, 0.999)$) for scene representation and camera tracking optimization. The learning rate for tracking on Replica, ScanNet, and TUM RGB-D dataset is $1\times 10^{-3}$, $5\times 10^{-4}$, $1\times 10^{-2}$. 

\subsection{Evaluation Datasets and Metrics}
We operate our system in different datasets ranging from small room scenes to large indoor scenes. We also collect our own dataset to evaluate the performance of our system in real-world scenarios and its deployment on mobile robots. To evaluate scene reconstruction results, we choose the Replica dataset\cite{straub2019replica}, which is a synthetic 3D indoor dataset from a room to an entire apartment scale. In order to create a more realistic depth input, we randomly remove the depth of some pixels and perturb the depth with Gaussian noise $\mathbb{N}(0,s^2)$, where the standard deviation increases with the depth value.  
\begin{figure*}[t]
    \centering
    \includegraphics[width=0.9\linewidth]{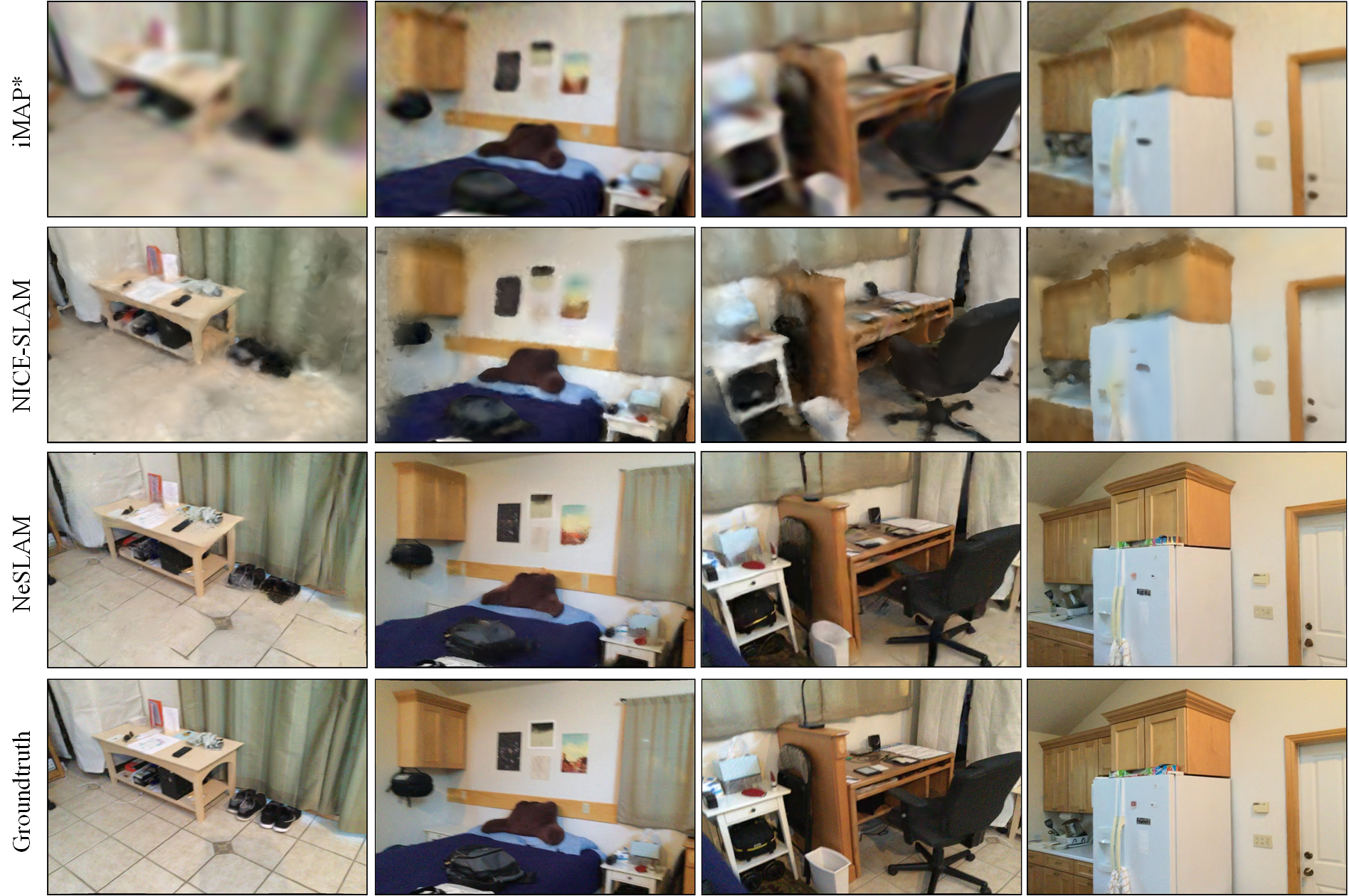}
    \vspace{-0.2cm}
    \caption{Qualitative results on the Scannet dataset~\cite{scannet}. We show the view synthesis results of iMAP*\cite{Sucar_2021_imap}, NICE-SLAM\cite{Zhu_2022_niceslam}, and our method. In large real-world indoor scenes, our method outperforms other algorithms in scene representation and view synthesis.}
    \label{fig:reconstruct1}
    \vspace{-0.4cm}
\end{figure*}
\begin{table}[]
\vspace{-0.2cm}
\centering
\caption{Comparision of Runtime in Replica dataset \cite{straub2019replica}.} 
\begin{tabular}{lll}
\toprule
Methods   & Tracking {[}s{]} & Mapping {[}s{]} \\ \midrule
iMAP~\cite{Sucar_2021_imap}      & 101.45   & 448.85  \\
NICE-SLAM~\cite{Zhu_2022_niceslam} & 47.88    & 140.74  \\
Ours      & \textbf{44.58}    & \textbf{130.58}  \\ \bottomrule
\end{tabular}
\vspace{-0.5cm}
\label{tab:runtime}
\end{table}
For camera tracking, we use TUM RGB-D dataset~\cite{tum}
to evaluate pose estimation with the given groundtruth trajectory. Moreover,
we consider ScanNet~\cite{scannet} to evaluate the scalability of our system.
Following \cite{Zhu_2022_niceslam,Sucar_2021_imap}, we evaluate Accuracy, Completion, Completion Ratio [$< 5cm\%$], and Depth L1 metrics for scene geometry. As for the evaluation of camera tracking results, we use Absolute Trajectory Error (ATE) Root Mean Squared Error (RMSE), Mean, and Median. We also use Peak Signal-to-noise Ratio (PSNR) to evaluate novel view synthesis results. Please note that iMAP* is the re-implementation of iMAP provided in \cite{Zhu_2022_niceslam}.

 \begin{table}[]
\centering
\caption{Camera tracking results on our own dataset and TUM RGB-D datasets~\cite{tum}. We use ATE RMSE [cm] as our evaluation metric. }
\begin{tabular}{lllll}
\toprule
Methods       & fr1/desk                       & fr2/xyz                        & fr3/office                     & ROOM \\ \hline
iMAP~\cite{Sucar_2021_imap}          & 4.93                           & 2.04                           & 5.84                           & 6.34 \\
NICE-SLAM~\cite{Zhu_2022_niceslam}     & 2.75                           & 1.83                           & 3.02                           & 4.73 \\
DI-Fusion~\cite{di-fusion}     & 4.45                           & 2.39                           & 15.73                          & 6.39 \\
Ours          & \textbf{1.83} & \textbf{1.09} & \textbf{2.14} & \textbf{2.95} \\ \midrule
BAD-SLAM~\cite{BADslam}      & 1.89                           & 1.21                           & 1.83                           & 2.88 \\
ElasticFusion~\cite{elasticfusion} & 2.04                           & 1.27                           & 1.71                           & 3.21 \\
ORB-SLAM2~\cite{orbslam2}     & \textbf{1.63}                           & \textbf{0.62}                           & \textbf{1.36}                           & 2.97 \\ \bottomrule
\end{tabular}
\label{tab:tum1}
\vspace{-0.4cm}
\end{table}

\begin{table}[!t]
\centering
\caption{Camera tracking results on the Scannet datasets~\cite{scannet}. We use ATE RMSE [cm] as our evaluation metric.}
\setlength{\tabcolsep}{1mm}{
\begin{tabular}{lccccccc}
\toprule
Scene ID & 0000          & 0059          & 0106          & 0169          & 0181          & 0207          & Avg.          \\ \midrule
iMAP*~\cite{Sucar_2021_imap}                       & 55.95         & 32.06         & 17.50         & 70.51         & 32.10         & 11.91         & 36.67         \\
NICE-SLAM~\cite{Zhu_2022_niceslam}                    & 8.64          & 12.25         & 8.09          & 10.28         & 12.93         & 5.59          & 9.63          \\
Ours                         & \textbf{6.87} & \textbf{7.37} & \textbf{5.23} & \textbf{9.07} & \textbf{9.27} & \textbf{4.08} & \textbf{6.98} \\ \bottomrule
\end{tabular}}
\label{tab:scannet}
\vspace{-0.5cm}
\end{table}

For Replica datasets\cite{straub2019replica}, it is a synthetic dataset. So, we use the processed RGB-D sequence with noisy depth input to better simulate real-world environments. The Gaussian noise is set to $\mathbb{N}(0,0.8)$. The quantitative and qualitative reconstruction results are shown in Table \ref{tab:reconstruct} and Fig.~\ref{fig:mesh_vis}. With the depth denoising and completion network and improved hierarchical scene representation method, our method can reconstruct the scene more precisely. In Fig.~\ref{fig:mesh_vis}, we can see that our algorithm significantly outperforms other algorithms in reconstruction accuracy, smoothness, and completeness. 
 To better showcase the reconstructed results, we have zoomed in on a specific region of the images. The left and right red boxes show the zoomed-in reconstruction results of the desk and the lamp, respectively. In Table~\ref{tab:reconstruct}, we can see that the accuracy metric is 21.4\% higher than NICE-SLAM. The improved hierarchical scene representation method with SDF value greatly enhances the capability of scene representation of our method. Our proposed depth denoising and completion algorithm also improves the reconstruction accuracy in the presence of noisy inputs, while other algorithms exhibit low accuracy when dealing with noisy inputs. 

\renewcommand\arraystretch{1.3}
\begin{table}[!t]
\centering
\caption{Ablation study on different datasets of different module. }
\scalebox{0.9}{
\setlength{\tabcolsep}{1mm}{
\begin{tabular}{lccccc}
\toprule
Methods                            &               & Replica & ScanNet & TUM RGB-D & ROOM  \\ \midrule
\multirow{4}{*}{(a)w/o $D_\theta$} & Acc.{[}cm{]} $\downarrow$ & 3.17    & -       & -         & -     \\
                                   & Depth{[}cm{]} $\downarrow$& 2.37    & 23.98   & 7.89      & 8.92  \\
                                   & RMSE{[}cm{]} $\downarrow$ & 0.73    & 7.93    & 1.82      & 3.29  \\
                                   & PSNR{[}db{]} $\uparrow$ & 25.56   & 22.80   & 21.84     & 21.76 \\ \midrule
\multirow{4}{*}{(b)w/o SDF}       & Acc.{[}cm{]} $\downarrow$ & 3.84    & -       & -         & -     \\
                                   & Depth{[}cm{]} $\downarrow$& 2.31    & 23.28   & 7.13      & 8.79  \\
                                   & RMSE{[}cm{]} $\downarrow$ & 0.83    & 8.87    & 2.08      & 3.47  \\
                                   & PSNR{[}db{]} $\uparrow$ & 25.43   & 22.43   & 21.23     & 21.18 \\ \midrule
\multirow{4}{*}{(c)w/o FT-Ref}     & Acc.{[}cm{]} $\downarrow$ & 3.01    & -       & -         & -     \\
                                   & Depth{[}cm{]} $\downarrow$& 2.03    & 23.19   & 6.99      & 8.68  \\
                                   & RMSE{[}cm{]} $\downarrow$ & 0.81    & 8.213   & 2.02      & 3.62  \\
                                   & PSNR{[}db{]}  & 25.75   & 22.51   & 22.01     & 21.57 \\ \midrule
\multirow{4}{*}{(d)w/o PW Loss}    & Acc.{[}cm{]} $\downarrow$ & 2.87    & -       & -         & -     \\
                                   & Depth{[}cm{]} $\downarrow$& 1.92    & 21.97   & 6.82      & 8.53  \\
                                   & RMSE{[}cm{]} $\downarrow$ & 0.74    & 7.82    & 1.83      & 3.24  \\
                                   & PSNR{[}db{]}  & 26.86   & 23.01   & 22.73     & 22.72 \\ \midrule
\multirow{4}{*}{(e)w/o ICP Loss}   & Acc.{[}cm{]} $\downarrow$ & 2.97    & -       & -         & -     \\
                                   & Depth{[}cm{]} & 1.98    & 21.82   & 6.93      & 8.75  \\
                                   & RMSE{[}cm{]}  & 0.73    & 8.03    & 1.98      & 3.44  \\
                                   & PSNR{[}db{]}  & 26.02   & 22.97   & 23.26     & 22.31 \\ \midrule
\multirow{4}{*}{NeSLAM (Full)}     & Acc.{[}cm{]}  & \textbf{2.57}    & -       & -         & -     \\
                                   & Depth{[}cm{]} & \textbf{1.90}    & \textbf{20.37}   & \textbf{6.75}      & \textbf{8.41}  \\
                                   & RMSE{[}cm{]}  & \textbf{0.69}    & \textbf{6.98}    & \textbf{1.68}      & \textbf{3.01}  \\
                                   & PSNR{[}db{]}  & \textbf{27.03}   & \textbf{23.88}   & \textbf{23.87}     & \textbf{23.59} \\ \bottomrule
\end{tabular}}}
\label{tab:ablation}
\vspace{-0.4cm}
\end{table}

 The camera tracking results are shown in Table~\ref{tab:tracking}. Our method outperforms all NeRF-based SLAM systems in all metrics. Compared with NICE-SLAM, our RMSE metric is 65.3\% higher on average, thanks to the NeRF-based self-supervised feature tracking method. Compared with the traditional SLAM system~\cite{orbslam2}, We can achieve competitive camera tracking performance, while providing dense and high-fidelity scene reconstruction performance. The dense mapping of the scene is really important in robots navigation and human interaction. The qualitative results of sampling points are shown in Fig.~\ref{fig:reconstruct6}. Our methods can effectively leverage environmental information for localization. The view synthesis results and depth estimation results are shown in Table~\ref{tab:photometric} and Fig.~\ref{fig:reconstruct}. Compared with iMAP\cite{Sucar_2021_imap} and NICE-SLAM\cite{Zhu_2022_niceslam}, Our depth L1 metric is \textbf{33\%} better than NICE-SLAM \cite{Zhu_2022_niceslam}. Our PSNR metric is \textbf{48\%} better than NICE-SLAM \cite{Zhu_2022_niceslam}.  Fig.~\ref{fig:reconstruct} provides the qualitative comparison of view synthesis between different methods. Our method achieves the most high-fidelity novel views results.
 
For our own real-world datasets, we collect data from two different scenes: a laboratory environment and a hospital ward scene. We use them to evaluate camera tracking performance and view synthesis in small room scenes with rapid camera movement, limited perspective, and relatively sparse view. As shown in Fig.~\ref{fig:reconstruct2}, we present our mobile robot platform equipped with a camera (Realsense D435i) and LiDAR (RS Lidar-16). We also present the scenario of our datasets. The room is equipped with the VICON motion capture system V2.2, which we use as ground truth for our dataset. We also use the lidar sensor to provide the groundtruth of our dataset. In Fig.~\ref{fig:reconstruct3}, we show the qualitative results of view synthesis in a real-world scenario (hospital). Our algorithm achieves better image synthesis results compared with other methods. In Table~\ref{tab:tum1}, we compare our method with other methods in real-world datasets. Our method performs better than iMAP \cite{Sucar_2021_imap}and NICE-SLAM \cite{Zhu_2022_niceslam} (with implicit representation) and reduces the gap between implicit SLAM with traditional SLAM (ORB-SLAM\cite{orbslam2}, BAD-SLAM\cite{BADslam}, ElasticFusion\cite{elasticfusion}). Due to the self-supervised keypoint detection, our system is more accurate and robust for different scenes. 
With the limitations in obtaining ground truth, we are unable to compare the localization accuracy in the hospital ward scene. So we only show the qualitative results in the hospital scene in Fig. \ref{fig:reconstruct4}. We also present our camera tracking results on the TUM RGB-D dataset~\cite{tum}.

For ScanNet datasets\cite{scannet}, we employ this dataset to evaluate the performance of our algorithm in large real-world indoor scenarios. We select different scenes to evaluate the scalability, camera tracking accuracy, and view synthesis results. As shown in Table \ref{tab:scannet}, compared with iMAP and NICE-SLAM, our method performs better in tracking accuracy. Our feature tracking algorithm provides more accurate and robust results in larger-scale scenes. In Fig.~\ref{fig:reconstruct1}, we show the qualitative results of view synthesis. Our mapping algorithm effectively addresses the issue of noise input in real-world environments. Our algorithm achieves the best image synthesis results with high clarity and completeness. 
\begin{figure}[t]
    \centering
    \includegraphics[width=0.95\linewidth]{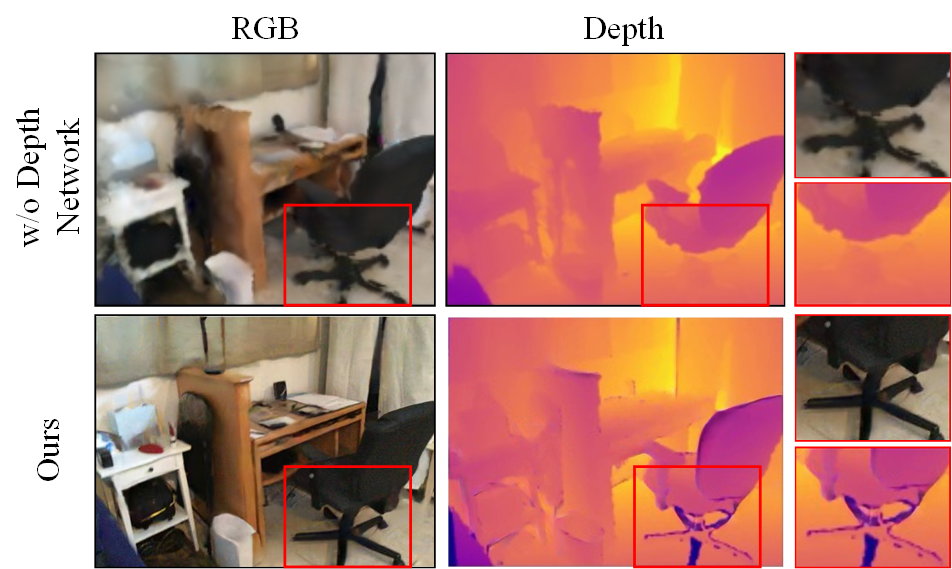}
    \caption{Qualitative results of our depth network ablation study on the Scannet dataset. We show the impact of the depth completion and denoising network on the final results.  }
    \label{fig:reconstruct5}
    \vspace{-0.4cm}
\end{figure}
We also compare the runtime for tracking and mapping. We modify our code to achieve better performance in time consumption. Thanks to the keypoint detection model, we can achieve better tracking performance and time consumption with fewer sampling pixels. We use 44 milliseconds for tracking and 147 milliseconds for mapping, compared with NICE-SLAM (50 milliseconds for tracking and 145 milliseconds for mapping). Our method is also robust to rapid camera movement and sudden frame loss. We provide extensive experiments in the supplementary material \textbf{https://github.com/dtc111111/NeSLAM}.
\vspace{-0.3cm}
\subsection{Ablation Study}
In this section, we conduct sufficient ablation studies to verify the effectiveness of our designed network. We show our ablation results in Table~\ref{tab:ablation}. (a) is NeSLAM without depth denoising and completion network. (b) is NeSLAM without SDF scene representation. (c) is NeSLAM without self-supervised feature tracking refinement (d) is NeSLAM without patch-wise loss (e) is NeSLAM without ICP loss. 

\noindent \textbf{Depth Denoising and Completion Network} \quad In Table \ref{tab:ablation} (a), we remove our designed depth network. It is obvious that this network has a great influence on depth L1 and PSNR metrics. This network significantly improves the capacity of scene geometry representation and enhances geometric consistency and robustness for noisy input. The qualitative results of our ablation study on depth network are shown in Fig.~\ref{fig:reconstruct3}. The depth network aids in recovering the geometric representation, ensuring geometric consistency across multi-view, and improving the results of depth estimation and view synthesis. \\
\noindent \textbf{Hierarchical Scene Representation with SDF} \quad In Table \ref{tab:ablation} (b), we replace the SDF hierarchical scene representation with original occupancy value. 
Our reconstruction and view synthesis metrics show a significant decrease. It indicates that the SDF transformation is really helpful in scene reconstruction.

\noindent \textbf{Feature Tracking Network} \quad
In Table \ref{tab:ablation} (c), we cancel the refinement of our self-supervised feature tracking network. We can see that the refinement network plays an important role in accurate camera tracking. It also makes our system more robust to rapid camera movement and sudden frame loss. 

\noindent \textbf{Loss Function Design} \quad As displayed in Table \ref{tab:ablation} (d), we use the original color and depth loss (without patch-wise loss). The reconstruction and tracking accuracy decreases, which verifies the effectiveness of this design. Table \ref{tab:ablation} (e) shows that the RMSE metric decreases greatly without the ICP loss. It is obvious that explicitly expressing the pose into loss function is effective for tracking.       
\vspace{-0.3cm}
\section{Conclusion}
This paper proposes a dense SLAM system NeSLAM, which combines neural implicit scene representation with the SLAM system. A depth denoising and completion network and a self-supervised feature tracking network are proposed. Our depth network provides dense depth images with depth uncertainty which can guide the neural point sampling and enhance scene geometry consistency. In addition, we incorporate the Signed Distance Field (SDF) value into the hierarchical feature grid, which can better represent scene geometry. Furthermore, the proposed NeRF-based self-supervised feature tracking network enables accurate camera tracking and enhances the robustness of our system. Our extensive experiments demonstrate the effectiveness and accuracy of our system in both scene reconstruction, tracking, and view synthesis in complex indoor scenes. In our future work, we will focus on dynamic scenes, aiming to achieve high reconstruction and localization accuracy.

\bibliographystyle{IEEEtran} 

\bibliography{egbib.bib}

\end{document}